\documentclass{article}

\usepackage{apacite}

\usepackage[final]{iai_neurips_2024}

\usepackage[utf8]{inputenc} 
\usepackage[T1]{fontenc}    
\usepackage{url}            
\usepackage{booktabs}       
\usepackage{amsfonts}       
\usepackage{nicefrac}       
\usepackage{microtype}      
\usepackage{xcolor}         
\usepackage{graphicx}

\usepackage{bbm}
\usepackage{enumitem}
\usepackage{amsfonts}
\usepackage{setspace}
\usepackage{amsmath}
\usepackage{url}
\usepackage{booktabs}
\usepackage{multirow}
\usepackage{multicol}
\usepackage{wrapfig}
\usepackage{caption}
\usepackage{subcaption}
\usepackage{soul}
\usepackage{dsfont}
\usepackage{rotating}
\usepackage{float}

\usepackage{xspace} 
\usepackage{tikz}
\usetikzlibrary{bayesnet}
\usetikzlibrary{arrows}
\usepackage{color}
\usetikzlibrary{backgrounds}

\usepackage{longtable}
\usepackage{caption} 
\usepackage{hyperref}       

\title{Exploiting Interpretable Capabilities with
Concept-Enhanced Diffusion and Prototype Networks}

\author{%
  Alba Carballo-Castro, Sonia Laguna, Moritz Vandenhirtz, Julia E. Vogt \\
  Department of Computer Science\\
  ETH Zürich\\
  Switzerland \\
}

\begin{document}

\maketitle

\begin{abstract}
Concept-based machine learning methods have increasingly gained importance due to the growing interest in making neural networks interpretable. However, concept annotations are generally challenging to obtain, making it crucial to leverage all their prior knowledge. By creating concept-enriched models that incorporate concept information into existing architectures, we exploit their interpretable capabilities to the fullest extent. In particular, we propose Concept-Guided Conditional Diffusion, which can generate visual representations of concepts, and Concept-Guided Prototype Networks, which can create a concept prototype dataset and leverage it to perform interpretable concept prediction. These results open up new lines of research by exploiting pre-existing information in the quest for rendering machine learning more human-understandable.

\end{abstract}


\section{Introduction}
\label{intro}


With an increasing number of decision-making processes relying on Machine Learning (ML) methods, the field of Interpretable ML has gained significant importance \citep{doshiRigorousScienceInterpretable2017, liptonMythosModelInterpretability2016} with concept-based methods being a prominent focus of the recent literature in this area. Concepts can be defined as variables that encode human-understandable information and can be used to provide explanations. Concept-based methods have been explored in prior work \citep{Lampert2009, Kumar2009}, with Concept Bottleneck Models (CBMs) \citep{Koh2020}, Concept Embedding Models (CEMs) \citep{espinosa2022concept} Concept Activation Vectors (CAVs) \citep{kim2018tcav}, and Concept Whitening \citep{Chen2020} among the most popular.





In this work, we explore how concept information can help increase interpretability beyond its current use by integrating it into pre-existing methods. We propose different means to leverage concept knowledge and obtain visual concept representations by introducing \textit{Concept-Guided Conditional Diffusion} and \textit{Concept-Guided Prototype Networks}. The former is based on diffusion models \citep{sohl-dickstein15, ho2020denoising}, particularly on conditional diffusion \citep{ho2021classifierfree}, to use the concept information for guidance and obtain visual representations of the concepts. The latter builds on already interpretable methods such as Prototypical Part Networks \citep{chen2019protopnet} to obtain prototypical patches (prototypes). They characterize concept information and allow for concept prediction and the creation of a \textit{concept prototype dataset}.


\textbf{Contributions} This work contributes to the line of research on concept-based methods in several ways by introducing concept-enhanced methods to leverage concept knowledge and exploit their interpretable capabilities. (\textit{i}) We present \textit{Concept-Guided Conditional Diffusion}, a generative method that incorporates concept knowledge to guide the generation of concept-based samples. (\textit{ii}) We introduce \textit{Concept-Guided Prototype Networks}, which allow us to (a) obtain interpretable concept predictions and (b) create a concept prototype dataset including a visual representation of the concepts. (\textit{iii}) We illustrate the success of these methods through real-world dataset applications.



\begin{figure}[]
\includegraphics[width=0.95\textwidth]{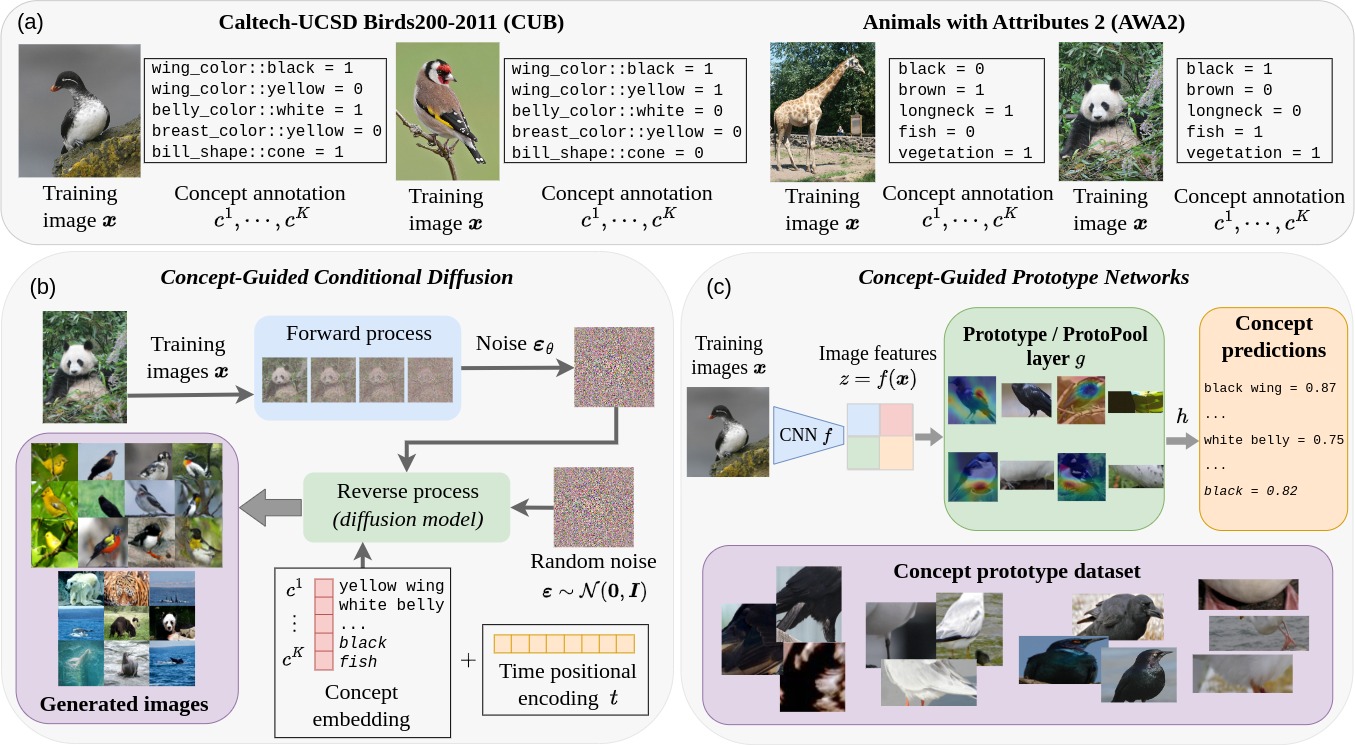}
\caption{Overview of the introduced concept-enhanced models. (a) Example of explored CUB and AWA2 datasets. \makebox{(b, c)} Summary of the methodology of Concept-Guided Conditional Diffusion and Prototype Networks. The resulting generations and prototype visualizations are shown in purple.}
\label{fig:schema}
\end{figure}

\section{Related Work}
\label{related_work}


\paragraph{Concept-based methods} Initially, concept-based methods aimed at attribute based classification~\citep{Lampert2009, Kumar2009}. Later on, Concept Activation Vectors (TCAV) \citep{kim2018tcav} and Invertible Concept-based Explanations (ICE)~\citep{Zhang2021ICE} were proposed as methods to interpret neural networks through human-understandable concepts. \citet{Koh2020} proposed Concept Bottleneck Models (CBM), which enforce an intermediate layer to correspond to concepts. Derivative works find ways to relax the bottleneck~\citep{espinosa2022concept}, don't require concept information during training~\citep{oikarinen2023labelfree, Yuksekgonul2023,marcinkevivcs2024beyond}, or improve the modeling capabilities~\citep{Sawada2022,Havasi2022, vandenhirtz2024stochastic}.
More recent works consider the concept bottleneck in the context of generative methods~\citep{Marconato2022,ismail2024concept}. Contrary to the restrictive enforcing of a concept bottleneck, our proposed generative method conditions on the complete concept information, allowing steerable generations of not only positive concept information (i.e., images containing a concept) but also negative, exploiting concept knowledge more exhaustively.

\paragraph{Diffusion models} Diffusion models were originally introduced by~\cite{sohl-dickstein15}, and have been followed by a series of works focusing on image generation and improved performance, such as noise conditional score network (NCSN)~\citep{song2019generative}, Denoising Diffusion Probabilistic Models (DDPM)~\citep{ho2020denoising} or Denoising Implicit Diffusion Models (DDIM)~\citep{song2022denoising}. Further works have dealt with Conditional Diffusion using a classifier as guidance  \citep{dhariwal2021diffusion} or without making use of it \citep{ho2021classifierfree}.

\paragraph{Prototype network} They perform classification by computing the similarity of the encoded input with a prototypical embedding~\citep{Li2018Prototypes}, which can be restricted to match an actual image patch~\citep{chen2019protopnet}. Other related works are ProtoPShare \citep{Rymarczyk2021ProtoPShare}, ProtoTrees \citep{nauta2021prototrees}, ProtoPool \citep{rymarczyk2022ProtoPool} and ProtoConcepts \citep{ma2023protoconcepts}, that learn prototypical concepts from the data rather than using a pre-annotated concept set.


\section{Methods}
\label{methods}



Concept-based methods are trained on triplets $\{(\boldsymbol{x}_i,y_i,\boldsymbol{c}_i\}_{i=1}^n$, where $\boldsymbol{x} \in \mathbb{R}^d$ are the inputs, $y \in \mathbb{R}$ is the target, and $\boldsymbol{c} \in \{0,1\}^k$ are different human-understandable concepts. Given $K$ concepts $\boldsymbol{c} = c^1, \ldots c^K$, we propose to incorporate this information into pre-existing methods by adapting the optimization problem to the multi-binary-label case, creating concept-enhanced models\footnote{The code is publicly available here: \href{https://github.com/acarballocastro/ConceptEnhanced}{\texttt{https://github.com/acarballocastro/ConceptEnhanced}}}. This way, we aim to acquire visual representations of concepts to enhance human interpretability. A summary of our methods is illustrated in Figure \ref{fig:schema}. In the following, we explore two ways to infuse pre-existing methods with concept information to obtain visual representations.


\subsection{Concept-Guided Conditional Diffusion}

Previous works on Conditional Diffusion use label information to guide the generations, but focus mostly on the multi-class setting \citep{dhariwal2021diffusion, ho2021classifierfree}. Our method extends classifier-free guidance \citep{ho2021classifierfree} to the multi-binary-label case to leverage concept knowledge, using the concept vector $\boldsymbol{c}$ to guide the diffusion process. The original approach parametrizes conditional and unconditional diffusion models through a single neural network. In the conditional model, label information is passed as an embedding. This label embedding is learned, and the row corresponding to the class label is added to the timestep positional encoding. Our method replaces the label information with a \textit{concept embedding} with K rows, one per concept.

Each datapoint $\boldsymbol{x}$ has per-concept binary annotations, representing its presence in the image. To allow for generated images to be guided by a subset of the available concepts, we introduce a user-defined binary mask $\boldsymbol{m}$ activated ($m^k = 1$) only for the concepts of interest. Given an embedding $\mathcal{E} = (\boldsymbol{e}_k)_{k=1}^K$, we select the rows of the mask-activated concepts in three different ways (we refer to Figure \ref{fig:embeddings} in Appendix \ref{app:CGDiff} for further details): \textbf{Positive Embedding}, where the selected embedding rows $\boldsymbol{e}_k$ are those activated by the mask $m_i^k = 1$ and with positive concept values, $c_i^k = 1$; \textbf{Opposite Embedding}, selects all rows activated by the mask $m_i^k = 1$ but inverting the values, $-\boldsymbol{e}_k$, in the case of negative concepts $c_i^k = 0$; and \textbf{Double Embedding}, which initializes two separate embeddings $\mathcal{E}_1$ and $\mathcal{E}_2$ and extracts the rows where $m_i^k = 1$ from $\mathcal{E}_1$ when $c_i^k = 1$, and $\mathcal{E}_2$ when $c_i^k = 0$.

Finally, the extracted rows are averaged and added to the positional encoding of the timesteps, used to guide the generations in the conditional model. Our proposed modeling allows us to introduce the guidance of generations not only with positive concepts $c_i^k = 1$ but also with negative concepts $c_i^k = 0$ (i.e., generations in which the concept of interest is \textit{not} present).

\subsection{Concept-Guided Prototype Networks}

In this section, we introduce the extension of ProtoPNet \citep{chen2019protopnet} and its subsequent work ProtoPools \citep{rymarczyk2022ProtoPool} to concept-enhanced methods by adapting them from the multi-class to the multi-binary label setting, allowing for interpretable concept prediction. In addition, prototypical image patches (prototypes) will be obtained for positive ($c_i^k = 1$) and negative ($c_i^k = 0$) concepts, resulting in $2 \times K$ concept classes for prototype training but just $K$ for concept prediction. 


\paragraph{Concept-Guided ProtoPNet} Given an input image $\boldsymbol{x}$, ProtoPNet \citep{chen2019protopnet} is composed by a CNN layer $f$ that extracts image features $\boldsymbol{z} = f(\boldsymbol{x})$, a prototype layer $g_{\boldsymbol{p}}$ to compute similarities between prototypes and patches of the convolutional output, and a fully connected layer $h$ for the final classification task. A summary of the original training algorithm is available in Appendix \ref{app:CGProto}.

If $\boldsymbol{P} = \{\boldsymbol{p}_j\}_{j=1}^m$ are the different $m$ prototypes, Concept-Guided ProtoPNet is built by modifying the loss function in Equation \ref{eq:ppnetloss}. The \textit{cross entropy} loss will be changed to the \textit{binary cross entropy} and the cluster (Clst) and separation (Sep) costs are adapted as 
\begin{equation}
    \text{Clst} = \frac{1}{n} \frac{1}{K} \sum_{i=1}^n \sum_{k=1}^K \min_{j: \boldsymbol{p}_j \in \boldsymbol{P}_k^{c_i^k}} \min_{\boldsymbol{z} \in \text{patches}(f(x_i))} \|\boldsymbol{z} - \boldsymbol{p}_j\|_2^2,
\end{equation}
and
\begin{equation}
    \text{Sep} = -\frac{1}{n} \frac{1}{K} \sum_{i=1}^n \sum_{k=1}^K \min_{j: \boldsymbol{p}_j \in \boldsymbol{P}_k^{\Bar{c}_i^k}} \min_{\boldsymbol{z} \in \text{patches}(f(x_i))} \|\boldsymbol{z} - \boldsymbol{p}_j\|_2^2,
\end{equation}
where $c_i^k$ is the binary value of the $k$-th concept for the $i$-th training sample and $\Bar{c}_i^k = 1 - c_i^k$.

Previously, Clst encouraged that there is at least one latent patch in every training image close to at least one prototype of its own class. Now, it will encourage having at least one latent patch in every training image close to at least one prototype of its concept class (which can be positive or negative) for all the different $K$ concepts. This is achieved by partitioning the initial prototype set $\boldsymbol{P} = \boldsymbol{P}_k^{c_i^k} \cup \boldsymbol{P}_k^{\Bar{c}_i^k}$ into the subsets of prototypes assigned to the concept classes of the given training image $x_i$ and those assigned to the opposite, respectively. Finally, Clst is calculated for all concepts belonging to the first subset and averaging across them. Similarly, Sep encouraged all patches from a training image to stay far from all the prototypes that are not of the same class. This was pertinent since labels were mutually exclusive, unlike now where each image has $K$ concepts associated. Therefore, Sep will encourage that, for a given training image and each of the $K$ concepts, there are no latent patches close to the prototypes assigned to the opposite concept classes. This will be achieved by calculating Sep for the subset $\boldsymbol{P}_k^{\Bar{c}_i^k}$ and averaging across all concepts.

For the prototype projection step, the update remains unchanged while accounting for the subsets of patches per concept being positive and negative. Finally, as the concepts are now not mutually exclusive, the weights of the last layer are initialized as $w_h^{(k,j)} = 1$ if $\boldsymbol{p}_j \in \boldsymbol{P}_k^1$, and $w_h^{(l,j)} = -1$ if $\boldsymbol{p}_j \notin \boldsymbol{P}_k^0$ for a given concept $k$. That is, the weights linking a positive concept logit with its prototypes are set to 1, whereas the ones linking the negative concept logits to its (negative) prototypes are set to $-1$. The rest of the weights are set to 0, encouraging a better classification. 


\paragraph{Concept-Guided ProtoPools} In ProtoPools \citep{rymarczyk2022ProtoPool}, prototypes can be shared across classes, and the allocation of prototypes to classes is dynamical instead of predetermined. To that end, the model uses prototype pool layer $g$, which learns a set of $m$ prototypes $\boldsymbol{P} = \{\boldsymbol{p}_j\}_{j=1}^m$ and a set of $S$ distributions $q_s \in \mathbb{R}^m$ per class. Each class has, therefore, $S$ \textit{slots}, and the distributions represent the probability of a prototype being assigned to one of the slots of that class. Instead of computing one similarity score per prototype unit, $g$ computes $S$ similarity scores per class which then pass by the fully connected layer to output the final prediction.

The main differences in the training algorithm with respect to ProtoPNet are the dynamical assignment of prototypes using a Gumbel-Softmax distribution and a new orthogonal loss to ensure that the same prototype is not assigned to more than one slot per class. The projection of prototypes and convex optimization of the last layer steps are also adapted (see Appendix \ref{app:CGProto} for further details).

To introduce Concept-Guided ProtoPools, we adapt it to the multi-binary-label case by calculating the cluster and separation costs as described before. In addition to the orthogonal loss in Equation \ref{eq:orthogonalwithin}, a second orthogonal loss is introduced to ensure the same prototype is not assigned to the positive and negative classes of a given concept:
\begin{equation}
    \text{Orth}_c = \sum_{i,j}^S \frac{\langle q_i, q_j \rangle}{\|q_i\|_2 \cdot \|q_j\|_2},
\end{equation}
where $i= 1, \ldots, S$ are the indexes of $q_i$ the distributions of concept $c^k$ and $j= 1, \ldots, S$ are the indexes of $q_j$ the distributions of the opposite concept $\Bar{c}^k$, with $k = 1, \ldots, K$.

The final loss function has the form:
\begin{equation}
\label{eq:CGPPoolLoss}
    \mathcal{L}_{CGPPool} = \frac{1}{n}\sum_{i=1}^n BCE \big[(h \circ g_{\boldsymbol{p}}  \circ f)(x_i), \ y_i\big] + \lambda_1 \text{Clst} + \lambda_2 \text{Sep} + \lambda_3 \text{Orth}_p + \lambda_4 \text{Orth}_c.
\end{equation}

For the projection of prototypes, $\mathcal{Z}_j$ stays the same as in Equation \ref{eq:ppoolproj}. Prototypes are assigned to the different $S$ slots of the $2 \times K$ concept classes, and then the model pushes each of them to the nearest training patch. Finally, for the convex optimization of the last layer, we set to 1 the weights that link a given positive concept to its $S$ slots and to $-1$ those that link a negative concept logit and its slots.

\section{Experimental setup}
\label{experiments}



Experiments were performed on the Caltech-UCSD Birds200-2011 dataset \citep{wah2011caltech}, in its adaptation from the CBM literature~\citep{Koh2020} with 112 concepts and 200 classes, and the Animals with Attributes 2 (AWA2) dataset \citep{xian2019awa2} comprising 85 concepts and 50 classes.

For the Concept-Guided Conditional Diffusion model, we performed experiments for the three types of embedding. After training, we sampled images with different combinations of concepts being positive and negative to later inspect the generations. As an example, for single-concept guided generations we chose \texttt{has\_wing\_color::black} and \texttt{has\_wing\_color::yellow}. This was further motivated as a means to explore the effect of the concept imbalance in the training images (41.4\% and 6.4\%, respectively). Generations with two concepts were made conditioning on \texttt{has\_wing\_color::black} and \texttt{has\_belly\_color::white}, and finally for three concepts on \texttt{has\_wing\_color::black}, \texttt{has\_belly\_color::white} and \texttt{has\_breast\_color::yellow}. In the case of the AwA2 dataset, selected concepts were \texttt{black} and \texttt{fish} (present in 60.9\% and 28.7\%), both for generations with one concept and with two concepts.


In Concept-Guided Prototype Networks, the goal is to show that these models can accurately predict the different binary concepts for a given input image in an interpretable way and to inspect the generated patches representative of each concept class. We will use the performance of a black-box pre-trained ResNet18 backbone used in the original CBM work~\citep{Koh2020} as an oracle for concept prediction. We measure the concept performance with the test set accuracy across varying architecture configurations, particularly DenseNets, ResNets, and VGGs. 

We generated ten prototypes per each of the $2 \times K$ concept classes in Concept-Guided ProtoPNet. As for Concept-Guided ProtoPools, we generated $m = 1000$ prototypes and assigned $S = 10$ slots per concept class. For the prototype datasets, we obtained the 50 closest patches to each prototype, resulting in 500 prototype images per concept class (in the case of ProtoPools, if a prototype is assigned more than once to the same concept class, the number of prototypes can be lower).


\section{Results}
\label{results}



\begin{figure}[ht]
    \centering
    \includegraphics[width=0.9\textwidth]{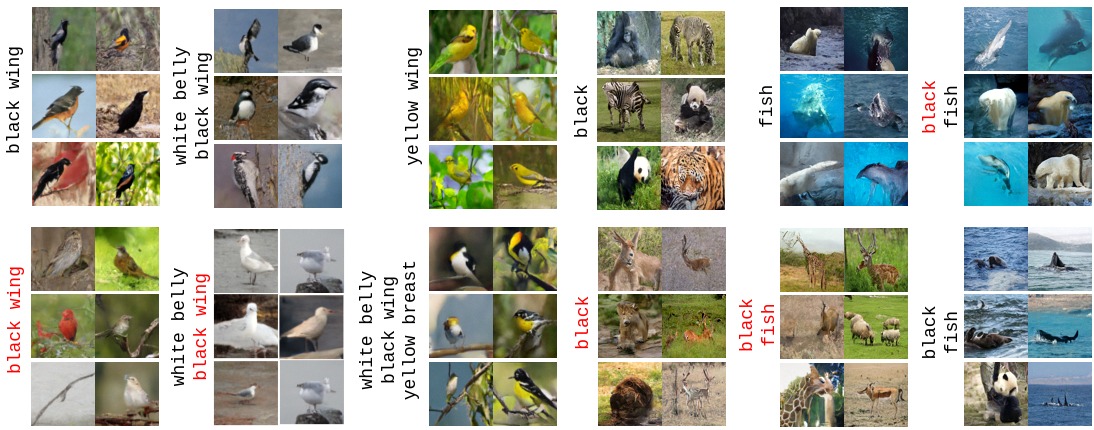}
    \caption{Concept visualizations (\texttt{positive} and \textcolor{red}{\texttt{negative}}) generated with Concept-Guided Conditional Diffusion for both datasets and different combinations of concepts. Each row corresponds to one of the embedding types described (positive, opposite, double).}
    \label{fig:results}
\end{figure}

\paragraph{Concept-Guided Conditional Diffusion} Figure \ref{fig:results} shows example generations for different concepts, complemented by the different sample images in Figure \ref{fig:schema} and a comprehensive overview in Appendix \ref{app:CUBCondDiff} (for CUB) and Appendix \ref{app:AwACondDiff} (for AWA2). It can be seen that the concepts of interest are present in all generations and, therefore, the model has been capable of generating visual concept representations with no remarkable differences between embedding types. In addition, generations also work for negative concepts, as it can be seen that the resulting images do not contain the concept when it is set to negative. In addition, we observe that the generations for concept \texttt{black wing} are better defined and have better quality than those of \texttt{yellow wing}, which is due to a higher proportion of images containing this concept. 

As for the images conditioned on different concepts at once, the concept combinations result in the desired generations for both positive and negative concepts. For instance, generations for concept \texttt{fish} positive and \texttt{black} negative result in images of animals that have fish in their diet but are not of color black. With this, we exploit the expressivity of concepts, particularly in the case where a direct human interpretation is not trivial. This can be later used as an intermediate step for a downstream task, for example, to increase interpretability in CBMs by generating visual representations of the latent concepts in the side channel.

\begin{figure}[htbp]
\centering
\begin{minipage}{0.42\textwidth}
    \centering
    \captionof{table}{Accuracy results for both Concept-Guided Prototype Networks with varying base architectures, \textbf{highlighting} the best-performing models. Results are compared with the black-box oracle.} 
    \label{tab:res-table}
    \begingroup

\setlength{\tabcolsep}{5pt} 
\renewcommand{\arraystretch}{1.1} 
\small
   \begin{tabular}{cc|cc}
\cline{2-4}
                     & Architecture            & \begin{tabular}[c]{@{}c@{}}Acc\\ (CUB)\end{tabular} & \begin{tabular}[c]{@{}c@{}}Acc \\ (AWA2)\end{tabular} \\ \cline{2-4} 
                     & \textit{Oracle} & \textit{0.961}                                           & \textit{0.901}                                             \\ \hline
\multirow{6}{*}{\rotatebox[origin=c]{90}{CG-PPNet}} & DenseNet121     & 0.874                                                    & 0.871                                                      \\
                     & DenseNet161     & 0.874                                                    & 0.855                                                      \\
                     & ResNet34        & \textbf{0.880}                                           & 0.852                                                      \\
                     & ResNet152       & 0.867                                                    & 0.842                                                      \\
                     & VGG16           & 0.870                                                    & \textbf{0.885}                                             \\
                     & VGG19           & 0.873                                                    & 0.879                                                      \\ \hline
                     
\multirow{4}{*}{\rotatebox[origin=c]{90}{CG-PPools}} & DenseNet121     & 0.867                                                    & 0.881                                                      \\
                     & DenseNet161     & 0.877                                                    & 0.883                                                      \\
                     & ResNet34        & \textbf{0.878}                                           & 0.877                                                      \\
                     & ResNet50        & 0.860                                                    & \textbf{0.892}                                             \\ \hline
\end{tabular}
\endgroup

\end{minipage}%
\hfill
\begin{minipage}{0.58\textwidth}
    \centering

\includegraphics[width=0.85\textwidth]{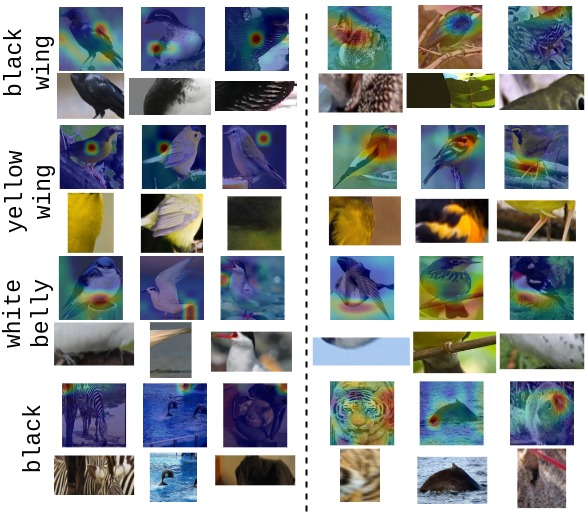}
\captionof{figure}{Concept prototype visualizations for Concept-Guided ProtoPNet (left) and ProtoPools (right). First row shows the activation map over the original image and the second row shows the corresponding prototype.}
\label{fig:results2}

\end{minipage}
\end{figure}

\paragraph{Concept-Guided Prototype Networks} Table \ref{tab:res-table} shows the main accuracy results for the different base architectures, complemented by further results in Tables \ref{ppnetaccuracy} (for ProtoPNet) and \ref{ppoolaccuracy} (for ProtoPools) in Appendix \ref{app:protoresults}. Despite a slight drop in concept accuracy with respect to the oracle, concept prototypes provide improved interpretability through a better understanding of the way models visualize concepts. In particular, AwA2 behaves on par with the oracle performance, especially in Concept-Guided ProtoPools, which highlights the method's success in bigger datasets.

Figure \ref{fig:results2} shows generated concept prototypes. The model can identify and extract relevant patches of the training images corresponding to the concept of interest. Figures \ref{fig:ProtoPPNetCUB}, \ref{fig:ProtoPPoolCUB}, \ref{fig:ProtoPPNetAwA} and \ref{fig:ProtoPPoolAwA} in Appendix~\ref{app:protoresults} display additional examples of prototypes alongside their closest patches for positive and negative concepts, calculated to conform the concept prototype dataset. Similarly, these methods allow to leverage concept information which can be used to increase downstream interpretability in other ML models. For instance, CBMs would benefit from having an interpretable concept predictor, which would help better understand and mitigate issues such as data leakage.

\section{Conclusion}
\label{conclusion}


%

In this paper, we introduced concept-enhanced methods as a way to incorporate concept information into existing architectures, and thus exploit their interpretable capabilities to the fullest extent. We proposed two different models \textit{Concept-Guided Conditional Diffusion} and \textit{Concept-Guided Prototype Networks}, which are able to generate concept-based samples or \textit{prototypes} and perform interpretable concept prediction. We applied these models to real-world datasets and showed successful image generation capabilities. This allows the visualization of concept information and an interpretable concept prediction without significant loss of performance with respect to black-box methods.
\paragraph{Limitations and future work} This work can be explored and impactful in the continuous concept domain. Another application lies in the context of CBMs, which are known to suffer from data leakage \citep{Mahinpei2021, Margeloiu2021, Havasi2022}, where the concept predictor conveys unintended information about the label. This issue could be further understood and mitigated with concept-enhanced methods, such as substituting usual black-box predictors for an interpretable one as a Concept-Guided Prototype Network; or calculating concept probabilities by similarity to concept prototypes or samples generated with Concept-Guided Conditional Diffusion. Lastly, both Concept-Guided methods could prove useful in obtaining visualizations to better understand the side channel. As a limitation, our method relies on concept-annotated datasets, which can be overcome in combination with current works on automated concept discovery \citep{Ghorbani2019, oikarinen2023labelfree}, generating images with diffusion models is computationally expensive and could be alleviated with recent less demanding variations \citep{rombach2021highresolution, guo2024makecheapscalingselfcascade}.

\begin{ack}
The project that gave rise to these results received the support of a fellowship from “la Caixa” Foundation (ID 100010434). The fellowship code is LCF/BQ/EU22/11930089. MV and SL are supported by the Swiss State Secretariat for Education, Research, and Innovation (SERI) under contract number MB22.00047.
\end{ack}


\bibliography{iai_neurips_2024}
\bibliographystyle{apacite}


\newpage
\appendix
\section{Methods}

In this section, we provide the reader with further details for a better understanding of the concept-enhanced methods implemented.

\subsection{Concept-Guided Conditional Diffusion}
\label{app:CGDiff}

\begin{figure}[ht]
\includegraphics[width=1\textwidth]{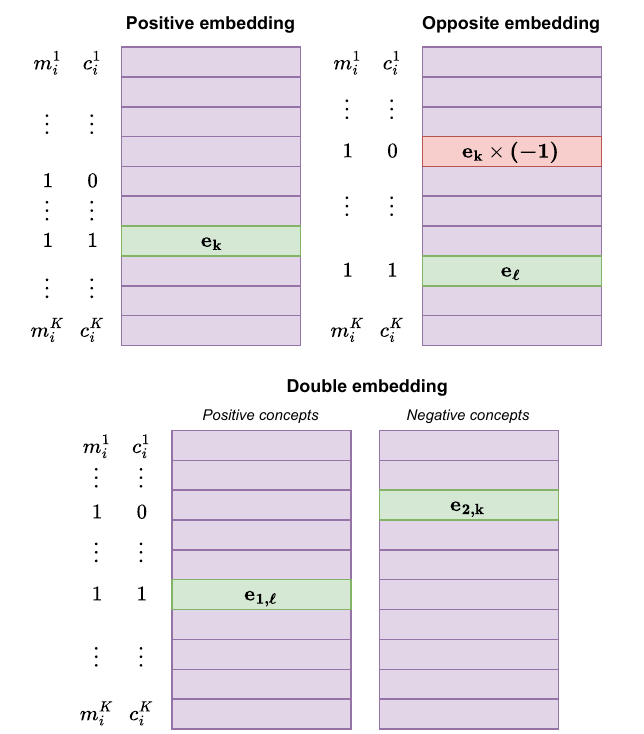}
\caption{Schematic representation of the three types of embedding proposed for the Concept-Guided Conditional Diffusion model.}
\label{fig:embeddings}
\end{figure}

Figure \ref{fig:embeddings} summarizes the three types of embedding for Concept-Guided Conditional Diffusion. In the \textit{positive embedding}, the situation in which a concept is not present for an image is treated similarly to when the concept is not relevant to guide the generations. In the \textit{opposite embedding}, the idea is inspired by methods from Natural Language Processing, where it is usual to force an embedding dimension to encode a semantic quality. Finally, for the \textit{double embedding}, it is ensured that positive and negative concepts are learned independently.

\subsection{Concept-Guided Prototype Networks}
\label{app:CGProto}

The steps to train a Prototypical Part Network (ProtoPNet) as originally described in \cite{chen2019protopnet} are the following:

\paragraph{(1) SDG of layers before the last layer} In this step, the model should learn the most representative patches of each class for the later classification of the images (i.e. the prototypes). The optimization problem is therefore posed on the convolutional layer parameters $w_{\text{conv}}$ and the different prototypes $\boldsymbol{P} = \{\boldsymbol{p}_j\}_{j=1}^m$ of $g_{\boldsymbol{p}}$. The aim is to minimize the loss:
\begin{equation}
\label{eq:ppnetloss}
    \mathcal{L}_{PPNet} = \frac{1}{n}\sum_{i=1}^n CrossEntropy \big[(h \circ g_{\boldsymbol{p}}  \circ f)(x_i), \ y_i\big] + \lambda_1 \text{Clst} + \lambda_2 \text{Sep},
\end{equation}
where
\begin{equation}
    \text{Clst} = \frac{1}{n} \sum_{i=1}^n \min_{j: \boldsymbol{p}_j \in \boldsymbol{P}_{y_i}} \min_{\boldsymbol{z} \in \text{patches}(f(x_i))} \|\boldsymbol{z} - \boldsymbol{p}_j\|_2^2,
\end{equation}
and
\begin{equation}
    \text{Sep} = -\frac{1}{n} \sum_{i=1}^n \min_{j: \boldsymbol{p}_j \notin \boldsymbol{P}_{y_i}} \min_{\boldsymbol{z} \in \text{patches}(f(x_i))} \|\boldsymbol{z} - \boldsymbol{p}_j\|_2^2.
\end{equation}

The first term of the loss, the cross entropy, penalizes incorrect classification of the training images. The cluster cost (Clst) encourages that there is at least one latent patch in every training image that is close to at least one prototype of its own class. Finally, the separation cost (Sep) encourages that all patches from a training image stay far from all the prototypes that are not of their same class.

\paragraph{(2) Projection of prototypes} To equate each of the prototypes that are in the latent space to actual patches of the training images, the model \textit{pushes} each prototype $\boldsymbol{p}_j$ to the nearest latent patch from the same class with the update
\begin{equation}
\label{eq:ppnettraining2}
    \boldsymbol{p}_j \leftarrow \arg \min_{\boldsymbol{z} \in \mathcal{Z}_j} \|\boldsymbol{z} - \boldsymbol{p}_j\|_2, \quad \mathcal{Z}_j = \{\Tilde{\boldsymbol{z}}: \Tilde{\boldsymbol{z}} \in \text{patches}(f(\boldsymbol{x}_i)), \forall i : y_i = l\},
\end{equation}
where $l = 1, \ldots, L$ are the different $L$ classes or labels.

\paragraph{(3) Convex optimization of the last layer} The weights of the last layer between the prototypes and the class logits are initialized as $w_h^{(l,j)} = 1$ if $\boldsymbol{p}_j \in \boldsymbol{P}_l$ and as $w_h^{(l,j)} = -0.5$ if $\boldsymbol{p}_j \notin \boldsymbol{P}_l$, where $\boldsymbol{P}_l \subseteq \boldsymbol{P}$ are the prototypes of class $l$. That is, depending on whether the weight connects the prototype of a given class with its class logit or not.

The authors propose a training step in which adjust the last layer connection weights $w_h^{(l,j)}$ to ensure sparsity by making $w_h^{(l,j)} \approx 0$ when $\boldsymbol{p}_j \notin \boldsymbol{P}_l$. According to the authors, this ensures that the model relies less on a negative reasoning process. Fixing all the parameters in the previous convolutional and prototype layers, the resulting convex optimization problem is
\begin{equation}
\label{eq:ppnetlastlayer}
    \min_{w_h} \frac{1}{n} \sum_{i=1}^n CrossEntropy \big[(h \circ g_{\boldsymbol{p}}  \circ f)(x_i), \ y_i\big] + \lambda_{last} \sum_{l=1}^L \sum_{j: \boldsymbol{p}_j \notin \boldsymbol{P}_l} |w_h^{(l,j)}|.
\end{equation}

The main differences to build ProtoPools \citep{rymarczyk2022ProtoPool} from ProtoPNet \citep{chen2019protopnet} are in the following steps:

\paragraph{Dynamical and differentiable assignment of prototypes} The authors suggest using the Gumbel-Softmax estimator \citep{jang2017categoricalreparameterizationgumbelsoftmax, maddison2017concretedistributioncontinuousrelaxation}   $\text{Gumbel-Softmax}(q_s, \tau) = (y_s^1, \ldots, y_s^m)$ where $\tau \in (0, \infty)$ is the temperature parameter,
\begin{equation}
    y_s^i = \frac{\exp\big((q_s^i + \eta_i) / \tau\big)}{\sum_{j=1}^m \exp\big((q_s^j + \eta_j) / \tau\big)},
\end{equation}
and $\eta_j$ are samples from the standard Gumbel distribution. This estimator generates distributions $q_s$ with exactly one of the probabilities close to 1, which results in a differentiable assignment of the prototypes to classes.

\paragraph{Orthogonal loss} The authors add an additional element to the loss function to ensure that the same prototype is not assigned to more than one slot per class,
\begin{equation}
\label{eq:orthogonalwithin}
    \text{Orth}_p = \sum_{i < j}^S \frac{\langle q_i, q_j \rangle}{\|q_i\|_2 \cdot \|q_j\|_2},
\end{equation}
which is calculated for each of the different classes.

\paragraph{Projection of prototypes} For the prototype projection, the idea is the same as in Equation \ref{eq:ppnettraining2}, with the difference that in this occasion
\begin{equation}
\label{eq:ppoolproj}
    \mathcal{Z}_j = \{\Tilde{\boldsymbol{z}}: \Tilde{\boldsymbol{z}} \in \text{patches}(f(\boldsymbol{x}_i)), \forall i : y_i \in L_j\},
\end{equation}
where $L_j$ is the set of classes such that one of its slots corresponds to prototype $\boldsymbol{p}_j$ and the assignment is made by means of the Gumbel-Softmax estimator.

\paragraph{Convex optimization of the last layer} In the adaptation to ProtoPools, the authors propose to initialize the weights between the different slots and their assigned class logit to 1, and the rest to 0.

\section{Experimental setup}

\paragraph{Concept-Guided Conditional Diffusion} Models were trained on the training split of the CUB dataset (4796 images) and on 40\% of the images of the AwA2 dataset ($sim$14292 images). All runs were trained
for 1500 epochs and for $T = 2000$ steps of diffusion. For the noise scheduler ($\beta_t$) we set a linear scheduler with $\beta_1 = 0.0001$ and $\beta_T = 0.2$. The learning rate was set at $0.0003$. Models were trained on an NVIDIA GeForce RTX 2080 Ti GPU to generate images of size $64 \times 64$. 

\paragraph{Concept-Guided ProtoPNet} Models were trained for 50 epochs each, with 5 epochs of warm-up and pushing and convex optimization of the last layer every 10 epochs. A summary of the different parameter combinations for the base encoder architecture, the prototype depth and the different coefficients of the loss is available in Table \ref{tab:PPNethyperparam}.
\begingroup
\setlength{\tabcolsep}{10pt} 
\renewcommand{\arraystretch}{1.25} 
\begin{table}[ht]
\centering
\caption{Hyperparameter combinations for Concept-Guided Prototype Network experiments.}
\label{tab:PPNethyperparam}
\begin{tabular}{c|c}
\hline
\textbf{Parameter}                                                                                           & \textbf{Values}                                                                                                                            \\ \hline
\begin{tabular}[c]{@{}c@{}}Base architecture \\ (prototype depth)\end{tabular}                               & \begin{tabular}[c]{@{}c@{}}VGG16 (128), VGG19 (128),\\ DenseNet121 (128), DenseNet161 (128),\\ ResNet34 (256), ResNet152 (512)\end{tabular} \\ \hline
\begin{tabular}[c]{@{}c@{}} ($\lambda_1, \lambda_2$)\end{tabular} & $(0.6, -0.06)$, $(0.8, -0.08)$, $(1, -0.1)$                                                                                                \\ \hline
\begin{tabular}[c]{@{}c@{}} $\lambda_{last}$\end{tabular}              & $10^{-3}$,  $10^{-4}$,  $10^{-5}$ \\
\hline
\end{tabular}%
\end{table}
\endgroup
\paragraph{Concept-Guided ProtoPool}  Models were trained for 100 epochs with 10 epochs of warm-up. Prototype depth was set as 256 and $\lambda_3 = \lambda_4 = 1$ for both orthogonal losses. The Gumbel-Softmax distribution was initialized with $\tau = 1$ and decreased for 30 epochs. The different hyperparameter combinations for base architecture and the rest of the loss coefficients are available in Table \ref{tab:PPoolhyperparam}.

\begingroup
\setlength{\tabcolsep}{10pt} 
\renewcommand{\arraystretch}{1.25} 
\begin{table}[ht]
\centering
\caption{Hyperparameter combinations for Concept-Guided ProtoPool experiments.}
\label{tab:PPoolhyperparam}
\begin{tabular}{c|c}
\hline
\textbf{Parameter}                                                                                           & \textbf{Values}                                                                                                                            \\ \hline
\begin{tabular}[c]{@{}c@{}}Base architecture \end{tabular}                               & \begin{tabular}[c]{@{}c@{}} DenseNet121, DenseNet161,\\ ResNet34, ResNet50\end{tabular} \\ \hline
\begin{tabular}[c]{@{}c@{}} ($\lambda_1, \lambda_2$)\end{tabular} & $(0.6, -0.06)$, $(0.8, -0.08)$, $(1, -0.1)$                                                                                                \\ \hline
\begin{tabular}[c]{@{}c@{}} $\lambda_{last}$\end{tabular}              & $10^{-4}$,  $10^{-5}$                            \\
\hline                                  
\end{tabular}%
\end{table}
\endgroup

\newpage

\section{Results for Concept-Guided Conditional Diffusion}

\subsection{Generated images for the CUB dataset}
\label{app:CUBCondDiff}

\begin{figure}[ht]
\includegraphics[width=1\textwidth]{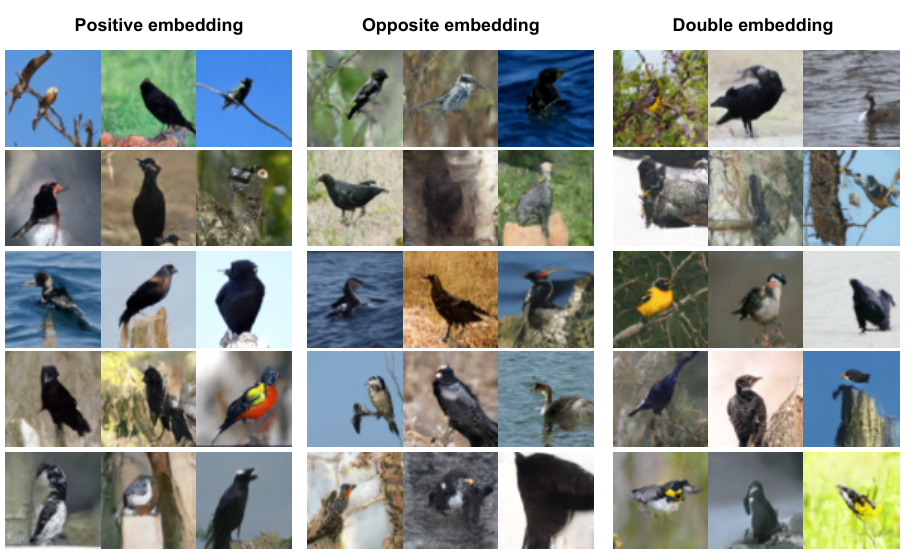}
\caption{Generations for concept \texttt{has\_wing\_color::black} positive.}
\label{fig:C8pos}
\end{figure}

\begin{figure}[ht]
\includegraphics[width=1\textwidth]{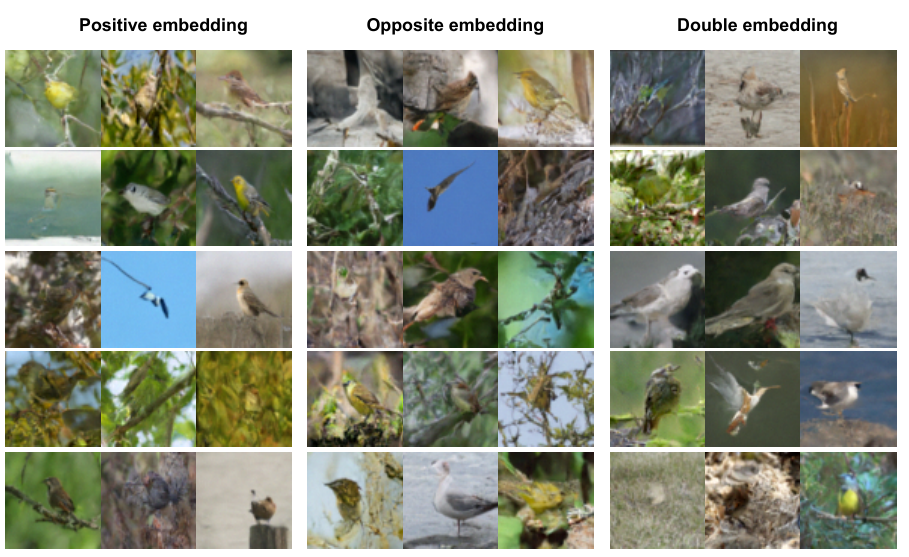}
\caption{Generations for concept \texttt{has\_wing\_color::black} negative.}
\label{fig:C8neg}
\end{figure}

\newpage

\begin{figure}[ht]
\centering
\includegraphics[width=1\textwidth]{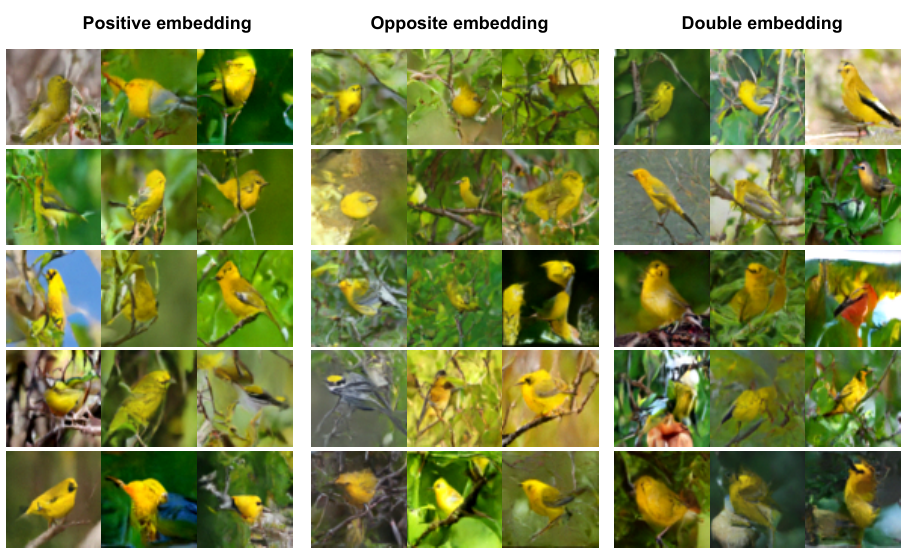}
\caption{Generations for concept \texttt{has\_wing\_color::yellow} positive.}
\label{fig:C7pos}
\end{figure}

\begin{figure}[ht]
\centering
\includegraphics[width=1\textwidth]{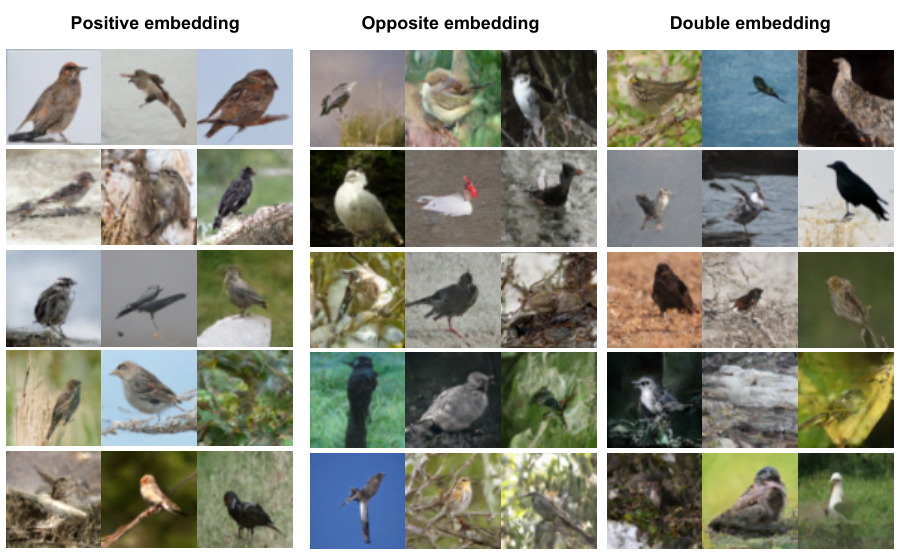}
\caption{Generations for concept \texttt{has\_wing\_color::yellow} negative.}
\label{fig:C7neg}
\end{figure}

\newpage

\begin{figure}[ht]
\centering
\includegraphics[width=1\textwidth]{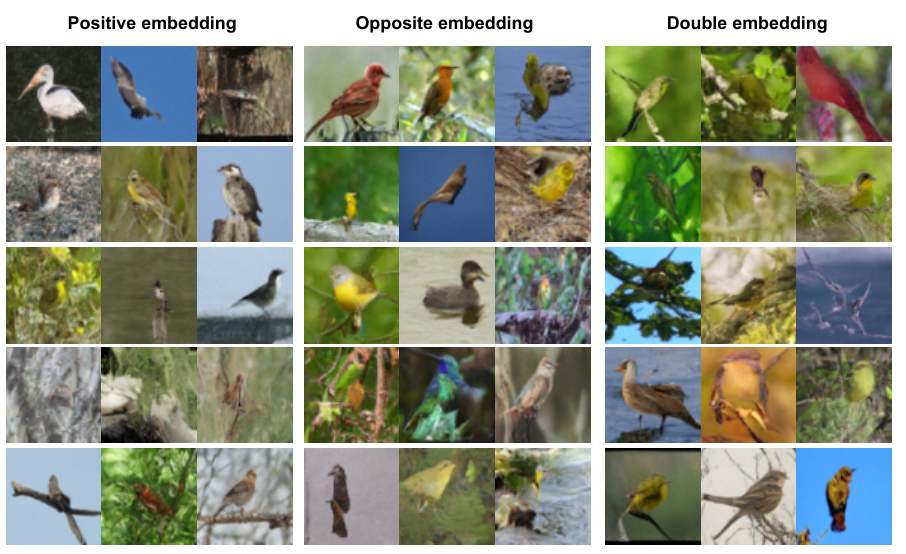}
\caption{Generations for \texttt{has\_wing\_color::black} and \texttt{has\_belly\_color::white} negative.}
\label{fig:C8_75none}
\end{figure}

\begin{figure}[ht]
\centering
\includegraphics[width=1\textwidth]{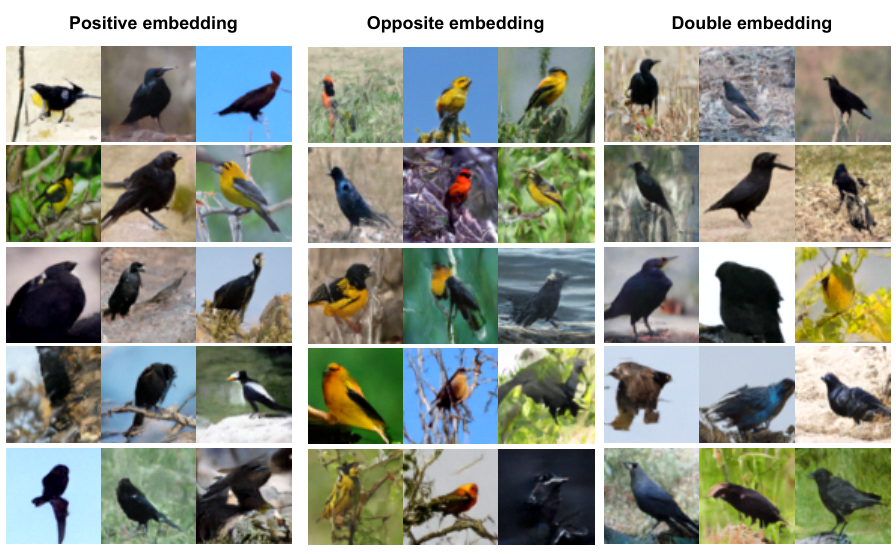}
\caption{Generations for concept \texttt{has\_wing\_color::black} positive and \texttt{has\_belly\_color::white} negative.}
\label{fig:C8_75black}
\end{figure}

\newpage

\begin{figure}[ht]
\centering
\includegraphics[width=1\textwidth]{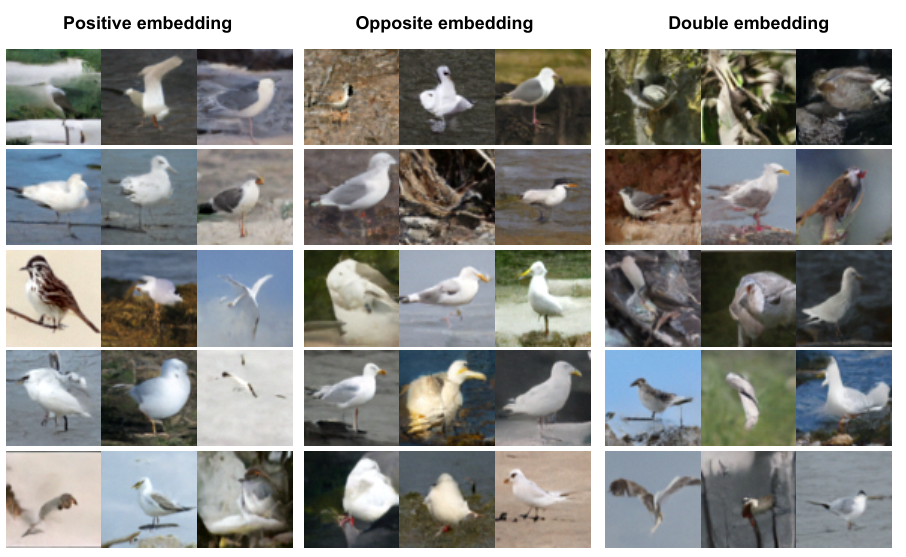}
\caption{Generations for concept \texttt{has\_belly\_color::white} positive and \texttt{has\_wing\_color::black} negative.}
\label{fig:C8_75neg}
\end{figure}

\begin{figure}[ht]
\centering
\includegraphics[width=1\textwidth]{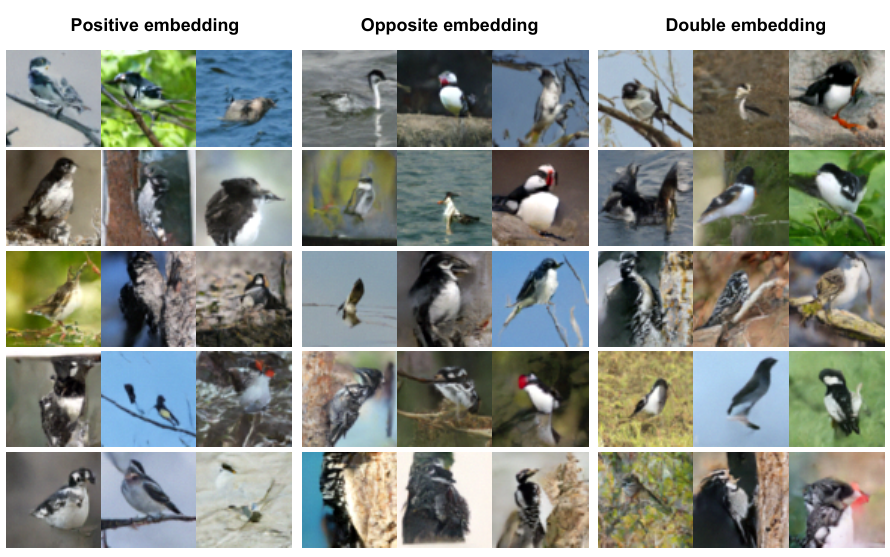}
\caption{Generations for \texttt{has\_wing\_color::black} and \texttt{has\_belly\_color::white} positive.}
\label{fig:C8_75both}
\end{figure}

\newpage

\begin{figure}[ht]
\centering
\includegraphics[width=0.95\textwidth]{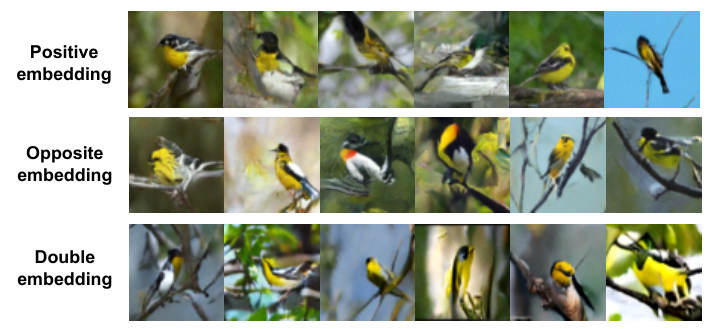}
\caption{Generations for concepts \texttt{has\_wing\_color::black}, \texttt{has\_belly\_color::white} and \texttt{has\_breast\_color::yellow}.}
\label{fig:Ctri_all}
\end{figure}
\vspace{2cm}
\begin{figure}[ht]
\centering
\includegraphics[width=1\textwidth]{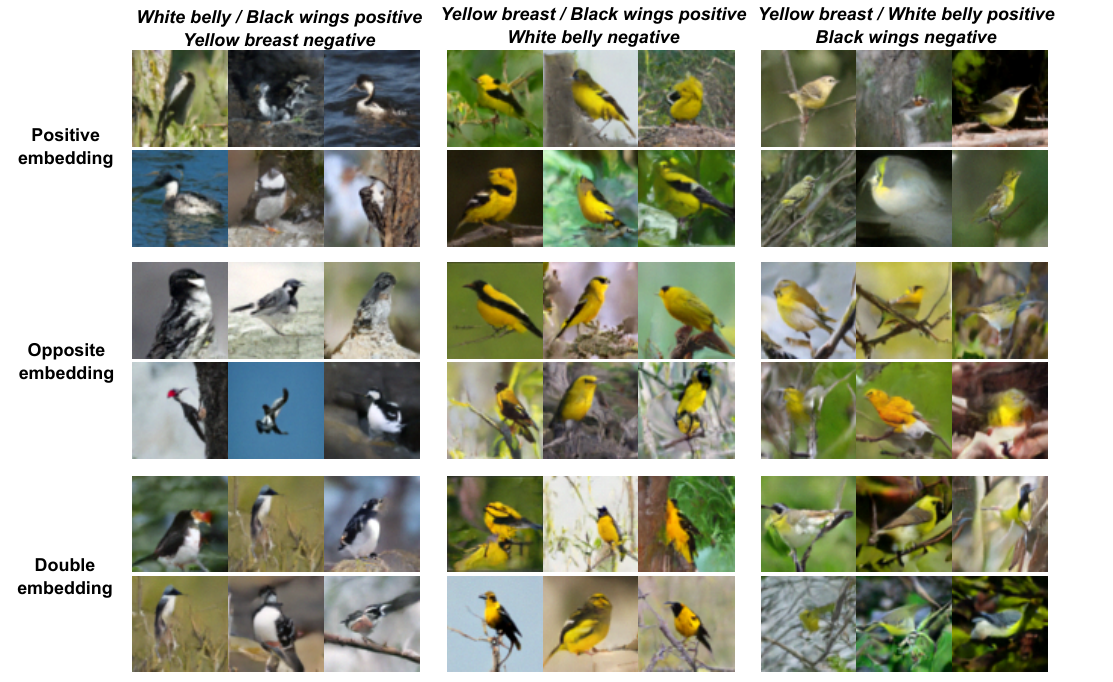}
\caption{Generations for different combinations of two concepts positive and one negative.}
\label{fig:Ctri_two}
\end{figure}

\newpage

\begin{figure}[h!]
\centering
\includegraphics[width=0.85\textwidth]{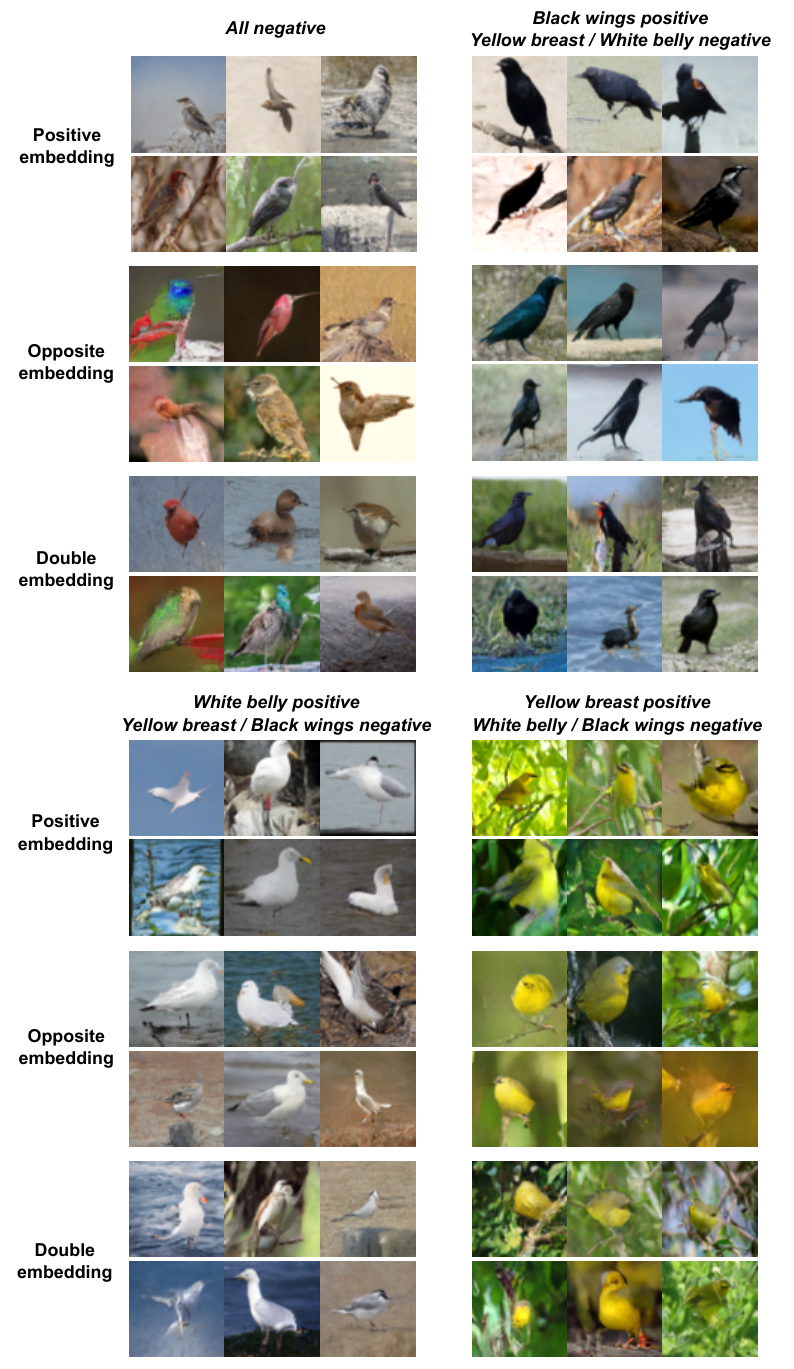}
\caption{Generations for none of the concepts or just one concept positive.}
\label{fig:Ctri_noneone}
\end{figure}

\newpage

\subsection{Generated images for the AwA2 dataset}
\label{app:AwACondDiff}
\vspace{-0.4cm}
\begin{figure}[h!]
\centering
\includegraphics[width=0.73\textwidth]{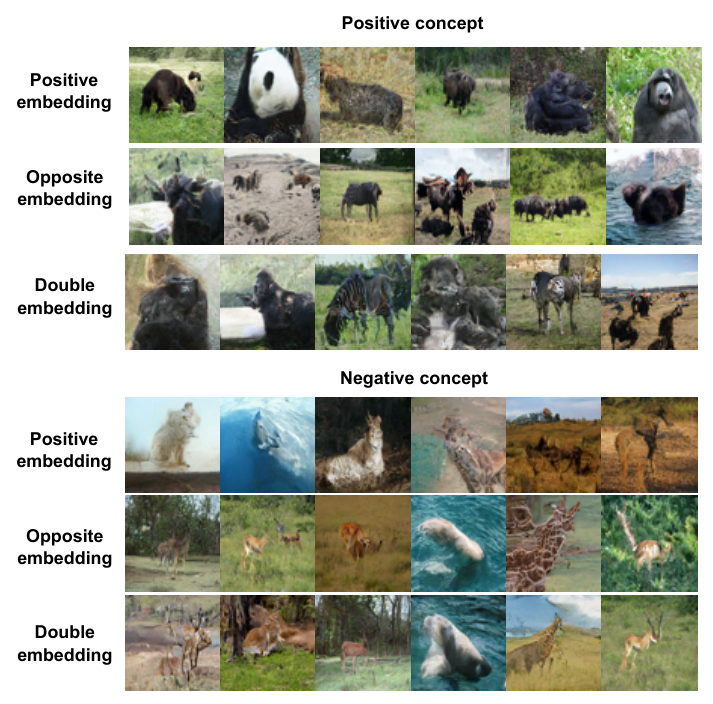}
\caption{Generations for concept \texttt{black} positive and negative.}
\label{fig:AwA_black}
\end{figure}
\vspace{-0.4cm}
\begin{figure}[h!]
\centering
\includegraphics[width=0.73\textwidth]{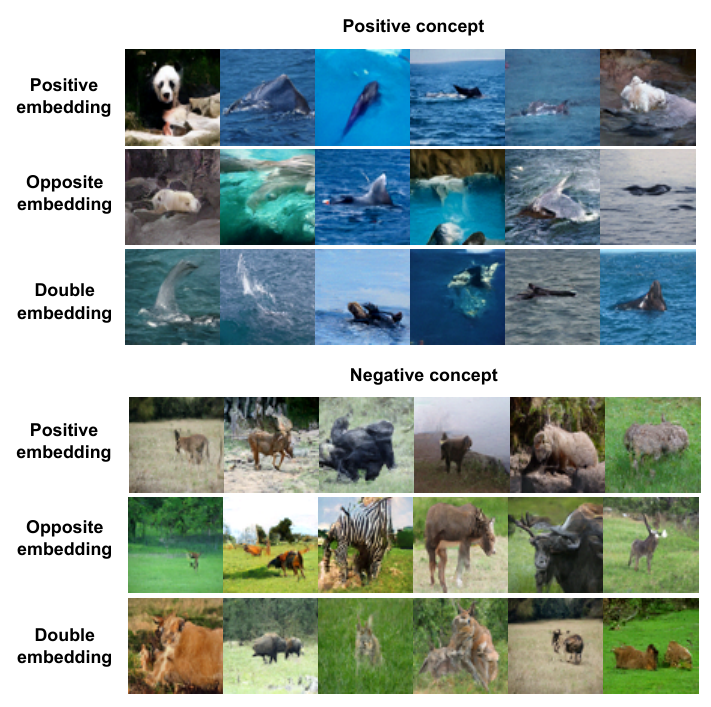}
\caption{Generations for concept \texttt{fish} positive and negative.}
\label{fig:AwA_fish}
\end{figure}
\newpage
\begin{figure}[h!]
\centering
\includegraphics[width=0.82\textwidth]{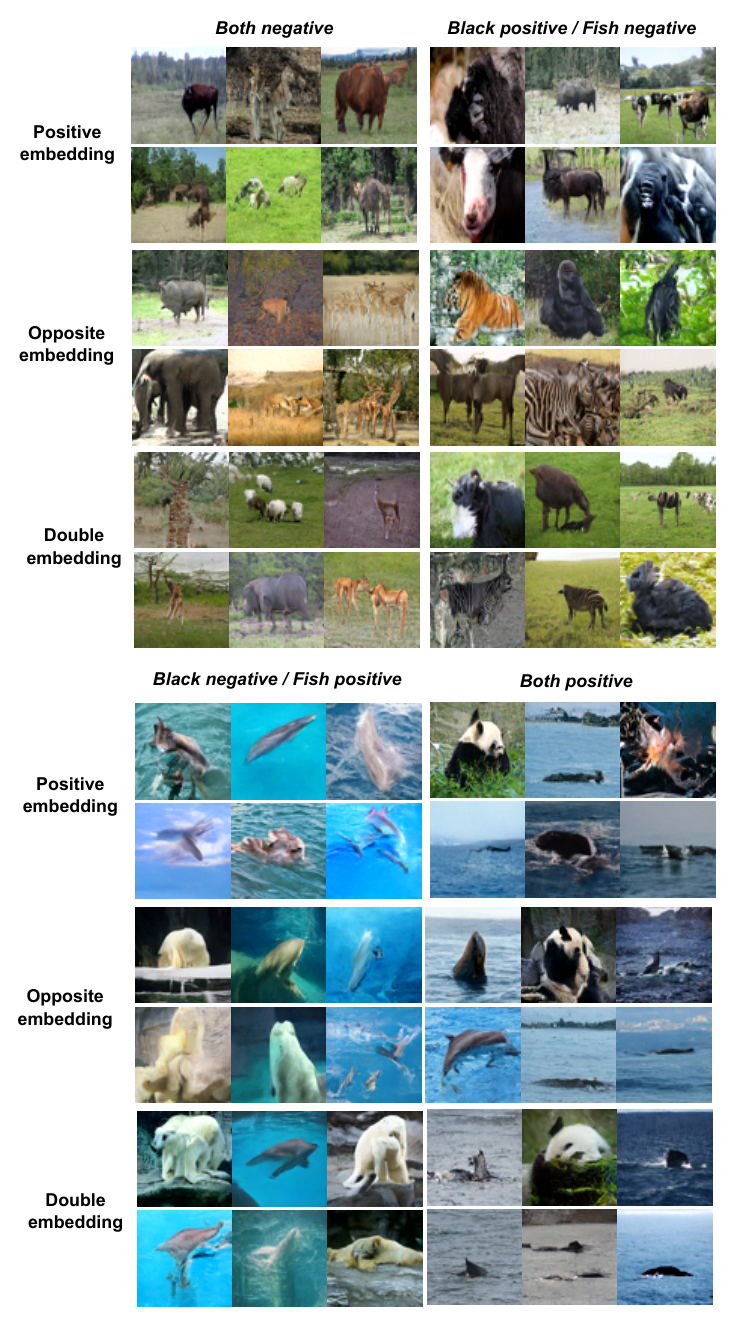}
\caption{Generations for different combinations of concepts \texttt{black} and \texttt{fish}.}
\label{fig:AwA_both}
\end{figure}

\newpage

\section{Results for Concept-Guided Prototype Networks}
\label{app:protoresults}
\begingroup
\setlength{\tabcolsep}{6pt} 
\renewcommand{\arraystretch}{1.13} 
\begin{longtable}{ccc|c|c}
\caption{Hyperparameter tuning accuracy in Concept-Guided ProtoPNet.} \label{ppnetaccuracy} \\

\hline
\multicolumn{1}{c|}{\textbf{\begin{tabular}[c]{@{}c@{}}Base architecture\\ (prototype depth)\end{tabular}}} & \multicolumn{1}{c|}{\textbf{$\lambda_{last}$}}  & \textbf{$\lambda_1$, $\lambda_2$} & \textbf{\begin{tabular}[c]{@{}c@{}}Accuracy\\ (CUB)\end{tabular}} & \textbf{\begin{tabular}[c]{@{}c@{}}Accuracy\\ (AwA2)\end{tabular}} \\ \hline
\multicolumn{3}{c|}{\textit{Oracle}}                                                                                                                                                              &    \textit{0.961}                                                               &                               \textit{0.901}                                     \\ \hline
\multicolumn{1}{c|}{\multirow{9}{*}{\begin{tabular}[c]{@{}c@{}}DenseNet121\\ (128)\end{tabular}}}           & \multicolumn{1}{c|}{$10^{-3}$}                  & (0.6, -0.06)                      & 0.864                                                             & 0.849                                                              \\
\multicolumn{1}{c|}{}                                                                                       & \multicolumn{1}{c|}{}                           & (0.8, -0.08)                      & 0.866                                                             & 0.836                                                              \\
\multicolumn{1}{c|}{}                                                                                       & \multicolumn{1}{c|}{}                           & (1, -0.1)                         & 0.860                                                             & 0.829                                                              \\ \cline{2-5} 
\multicolumn{1}{c|}{}                                                                                       & \multicolumn{1}{c|}{\multirow{3}{*}{$10^{-4}$}} & (0.6, -0.06)                      & 0.868                                                             & 0.855                                                              \\
\multicolumn{1}{c|}{}                                                                                       & \multicolumn{1}{c|}{}                           & (0.8, -0.08)                      & 0.866                                                             & 0.855                                                              \\
\multicolumn{1}{c|}{}                                                                                       & \multicolumn{1}{c|}{}                           & (1, -0.1)                         & 0.865                                                             & 0.854                                                              \\ \cline{2-5} 
\multicolumn{1}{c|}{}                                                                                       & \multicolumn{1}{c|}{\multirow{3}{*}{$10^{-5}$}} & (0.6, -0.06)                      & 0.874                                                             & 0.871                                                              \\
\multicolumn{1}{c|}{}                                                                                       & \multicolumn{1}{c|}{}                           & (0.8, -0.08)                      & 0.874                                                             & 0.867                                                              \\
\multicolumn{1}{c|}{}                                                                                       & \multicolumn{1}{c|}{}                           & (1, -0.1)                         & 0.872                                                             & 0.838                                                              \\ \hline
\multicolumn{1}{c|}{\multirow{9}{*}{\begin{tabular}[c]{@{}c@{}}DenseNet161\\ (128)\end{tabular}}}           & \multicolumn{1}{c|}{\multirow{3}{*}{$10^{-3}$}} & (0.6, -0.06)                      & 0.859                                                             & 0.833                                                              \\
\multicolumn{1}{c|}{}                                                                                       & \multicolumn{1}{c|}{}                           & (0.8, -0.08)                      & 0.859                                                             & 0.830                                                              \\
\multicolumn{1}{c|}{}                                                                                       & \multicolumn{1}{c|}{}                           & (1, -0.1)                         & 0.845                                                             & 0.813                                                              \\ \cline{2-5} 
\multicolumn{1}{c|}{}                                                                                       & \multicolumn{1}{c|}{\multirow{3}{*}{$10^{-4}$}} & (0.6, -0.06)                      & 0.865                                                             & 0.835                                                              \\
\multicolumn{1}{c|}{}                                                                                       & \multicolumn{1}{c|}{}                           & (0.8, -0.08)                      & 0.865                                                             & 0.836                                                              \\
\multicolumn{1}{c|}{}                                                                                       & \multicolumn{1}{c|}{}                           & (1, -0.1)                         & 0.855                                                             & 0.820                                                              \\ \cline{2-5} 
\multicolumn{1}{c|}{}                                                                                       & \multicolumn{1}{c|}{\multirow{3}{*}{$10^{-5}$}} & (0.6, -0.06)                      & 0.874                                                             & 0.855                                                              \\
\multicolumn{1}{c|}{}                                                                                       & \multicolumn{1}{c|}{}                           & (0.8, -0.08)                      & 0.873                                                             & 0.847                                                              \\
\multicolumn{1}{c|}{}                                                                                       & \multicolumn{1}{c|}{}                           & (1, -0.1)                         & 0.867                                                             & 0.847                                                              \\ \hline
\multicolumn{1}{c|}{\multirow{9}{*}{\begin{tabular}[c]{@{}c@{}}ResNet 34\\ (256)\end{tabular}}}             & \multicolumn{1}{c|}{\multirow{3}{*}{$10^{-3}$}} & (0.6, -0.06)                      & 0.869                                                             & 0.837                                                              \\
\multicolumn{1}{c|}{}                                                                                       & \multicolumn{1}{c|}{}                           & (0.8, -0.08)                      & 0.864                                                             & 0.826                                                              \\
\multicolumn{1}{c|}{}                                                                                       & \multicolumn{1}{c|}{}                           & (1, -0.1)                         & 0.854                                                             & 0.819                                                              \\ \cline{2-5} 
\multicolumn{1}{c|}{}                                                                                       & \multicolumn{1}{c|}{\multirow{3}{*}{$10^{-4}$}} & (0.6, -0.06)                      & 0.875                                                             & 0.840                                                              \\
\multicolumn{1}{c|}{}                                                                                       & \multicolumn{1}{c|}{}                           & (0.8, -0.08)                      & 0.870                                                             & 0.825                                                              \\
\multicolumn{1}{c|}{}                                                                                       & \multicolumn{1}{c|}{}                           & (1, -0.1)                         & 0.869                                                             & 0.832                                                              \\ \cline{2-5} 
\multicolumn{1}{c|}{}                                                                                       & \multicolumn{1}{c|}{\multirow{3}{*}{$10^{-5}$}} & (0.6, -0.06)                      & 0.878                                                             & 0.852                                                              \\
\multicolumn{1}{c|}{}                                                                                       & \multicolumn{1}{c|}{}                           & (0.8, -0.08)                      & 0.880                                                             & 0.844                                                              \\
\multicolumn{1}{c|}{}                                                                                       & \multicolumn{1}{c|}{}                           & (1, -0.1)                         & 0.878                                                             & 0.825                                                              \\ \hline
\multicolumn{1}{c|}{\multirow{9}{*}{\begin{tabular}[c]{@{}c@{}}ResNet152\\ (512)\end{tabular}}}             & \multicolumn{1}{c|}{\multirow{3}{*}{$10^{-3}$}} & (0.6, -0.06)                      & 0.842                                                             & 0.812                                                              \\
\multicolumn{1}{c|}{}                                                                                       & \multicolumn{1}{c|}{}                           & (0.8, -0.08)                      & 0.845                                                             & 0.818                                                              \\
\multicolumn{1}{c|}{}                                                                                       & \multicolumn{1}{c|}{}                           & (1, -0.1)                         & 0.834                                                             & 0.812                                                              \\ \cline{2-5} 
\multicolumn{1}{c|}{}                                                                                       & \multicolumn{1}{c|}{\multirow{3}{*}{$10^{-4}$}} & (0.6, -0.06)                      & 0.857                                                             & 0.809                                                              \\
\multicolumn{1}{c|}{}                                                                                       & \multicolumn{1}{c|}{}                           & (0.8, -0.08)                      & 0.859                                                             & 0.830                                                              \\
\multicolumn{1}{c|}{}                                                                                       & \multicolumn{1}{c|}{}                           & (1, -0.1)                         & 0.863                                                             & 0.809                                                              \\ \cline{2-5} 
\multicolumn{1}{c|}{}                                                                                       & \multicolumn{1}{c|}{\multirow{3}{*}{$10^{-5}$}} & (0.6, -0.06)                      & 0.867                                                             & 0.825                                                              \\
\multicolumn{1}{c|}{}                                                                                       & \multicolumn{1}{c|}{}                           & (0.8, -0.08)                      & 0.857                                                             & 0.842                                                              \\
\multicolumn{1}{c|}{}                                                                                       & \multicolumn{1}{c|}{}                           & (1, -0.1)                         & 0.848                                                             & 0.825                                                              \\ \hline
\multicolumn{1}{c|}{\multirow{9}{*}{\begin{tabular}[c]{@{}c@{}}VGG16\\ (128)\end{tabular}}}                 & \multicolumn{1}{c|}{\multirow{3}{*}{$10^{-3}$}} & (0.6, -0.06)                      & 0.857                                                             & 0.858                                                              \\
\multicolumn{1}{c|}{}                                                                                       & \multicolumn{1}{c|}{}                           & (0.8, -0.08)                      & 0.851                                                             & 0.849                                                              \\
\multicolumn{1}{c|}{}                                                                                       & \multicolumn{1}{c|}{}                           & (1, -0.1)                         & 0.852                                                             & 0.850                                                              \\ \cline{2-5} 
\multicolumn{1}{c|}{}                                                                                       & \multicolumn{1}{c|}{\multirow{3}{*}{$10^{-4}$}} & (0.6, -0.06)                      & 0.865                                                             & 0.873                                                              \\
\multicolumn{1}{c|}{}                                                                                       & \multicolumn{1}{c|}{}                           & (0.8, -0.08)                      & 0.855                                                             & 0.869                                                              \\
\multicolumn{1}{c|}{}                                                                                       & \multicolumn{1}{c|}{}                           & (1, -0.1)                         & 0.860                                                             & 0.863                                                              \\ \cline{2-5} 
\multicolumn{1}{c|}{}                                                                                       & \multicolumn{1}{c|}{\multirow{3}{*}{$10^{-5}$}} & (0.6, -0.06)                      & 0.870                                                             & 0.885                                                              \\
\multicolumn{1}{c|}{}                                                                                       & \multicolumn{1}{c|}{}                           & (0.8, -0.08)                      & 0.870                                                             & 0.875                                                              \\
\multicolumn{1}{c|}{}                                                                                       & \multicolumn{1}{c|}{}                           & (1, -0.1)                         & 0.869                                                             & 0.867                                                              \\ \hline
\multicolumn{1}{c|}{\multirow{9}{*}{\begin{tabular}[c]{@{}c@{}}VGG19\\ (128)\end{tabular}}}                 & \multicolumn{1}{c|}{\multirow{3}{*}{$10^{-3}$}} & (0.6, -0.06)                      & 0.854                                                             & 0.852                                                              \\
\multicolumn{1}{c|}{}                                                                                       & \multicolumn{1}{c|}{}                           & (0.8, -0.08)                      & 0.852                                                             & 0.851                                                              \\
\multicolumn{1}{c|}{}                                                                                       & \multicolumn{1}{c|}{}                           & (1, -0.1)                         & 0.847                                                             & 0.850                                                              \\ \cline{2-5} 
\multicolumn{1}{c|}{}                                                                                       & \multicolumn{1}{c|}{\multirow{3}{*}{$10^{-4}$}} & (0.6, -0.06)                      & 0.865                                                             & 0.879                                                              \\
\multicolumn{1}{c|}{}                                                                                       & \multicolumn{1}{c|}{}                           & (0.8, -0.08)                      & 0.860                                                             & 0.865                                                              \\
\multicolumn{1}{c|}{}                                                                                       & \multicolumn{1}{c|}{}                           & (1, -0.1)                         & 0.852                                                             & 0.848                                                              \\ \cline{2-5} 
\multicolumn{1}{c|}{}                                                                                       & \multicolumn{1}{c|}{\multirow{3}{*}{$10^{-5}$}} & (0.6, -0.06)                      & 0.871                                                             & 0.867                                                              \\
\multicolumn{1}{c|}{}                                                                                       & \multicolumn{1}{c|}{}                           & (0.8, -0.08)                      & 0.873                                                             & 0.870                                                              \\
\multicolumn{1}{c|}{}                                                                                       & \multicolumn{1}{c|}{}                           & (1, -0.1)                         & 0.868                                                             & 0.867                                                              \\ \hline
\end{longtable}
\endgroup

\begingroup
\setlength{\tabcolsep}{6pt} 
\renewcommand{\arraystretch}{1.3} 
\begin{longtable}{ccc|c|c}
\caption{Hyperparameter tuning accuracy in Concept-Guided ProtoPool.} \label{ppoolaccuracy} \\
\hline
\multicolumn{1}{c|}{\textbf{\begin{tabular}[c]{@{}c@{}}Base architecture\\ (prototype depth)\end{tabular}}} & \multicolumn{1}{c|}{\textbf{$\lambda_{last}$}}  & \textbf{$\lambda_1$, $\lambda_2$} & \textbf{\begin{tabular}[c]{@{}c@{}}Accuracy\\ (CUB)\end{tabular}} & \textbf{\begin{tabular}[c]{@{}c@{}}Accuracy\\ (AwA2)\end{tabular}} \\ \hline
\multicolumn{3}{c|}{\textit{Oracle}}                                                                                                                                                                         & \multicolumn{1}{c|}{\textit{0.961}}   & \multicolumn{1}{c}{\textit{0.901}}     \\ \hline
\multicolumn{1}{c|}{\multirow{6}{*}{DenseNet121}}                                                           & \multicolumn{1}{c|}{\multirow{3}{*}{$10^{-4}$}} & (0.6, -0.06)                        & 0.817                   & 0.879                    \\
\multicolumn{1}{c|}{}                                                                                       & \multicolumn{1}{c|}{}                           & (0.8, -0.08)                        & 0.816                   & 0.876                    \\
\multicolumn{1}{c|}{}                                                                                       & \multicolumn{1}{c|}{}                           & (1, -0.1)                           & 0.814                   & 0.875                    \\ \cline{2-5} 
\multicolumn{1}{c|}{}                                                                                       & \multicolumn{1}{c|}{\multirow{3}{*}{$10^{-5}$}} & (0.6, -0.06)                        & 0.867                   & 0.881                    \\
\multicolumn{1}{c|}{}                                                                                       & \multicolumn{1}{c|}{}                           & (0.8, -0.08)                        & 0.860                   & 0.877                    \\
\multicolumn{1}{c|}{}                                                                                       & \multicolumn{1}{c|}{}                           & (1, -0.1)                           & 0.854                   & 0.875                    \\ \hline
\multicolumn{1}{c|}{\multirow{6}{*}{DenseNet161}}                                                           & \multicolumn{1}{c|}{\multirow{3}{*}{$10^{-4}$}} & (0.6, -0.06)                        & 0.815                   & 0.882                    \\
\multicolumn{1}{c|}{}                                                                                       & \multicolumn{1}{c|}{}                           & (0.8, -0.08)                        & 0.816                   & 0.880                    \\
\multicolumn{1}{c|}{}                                                                                       & \multicolumn{1}{c|}{}                           & (1, -0.1)                           & 0.815                   & 0.883                    \\ \cline{2-5} 
\multicolumn{1}{c|}{}                                                                                       & \multicolumn{1}{c|}{\multirow{3}{*}{$10^{-5}$}} & (0.6, -0.06)                        & 0.870                   & 0.878                    \\
\multicolumn{1}{c|}{}                                                                                       & \multicolumn{1}{c|}{}                           & (0.8, -0.08)                        & 0.877                   & 0.878                    \\
\multicolumn{1}{c|}{}                                                                                       & \multicolumn{1}{c|}{}                           & (1, -0.1)                           & 0.873                   & 0.880                    \\ \hline
\multicolumn{1}{c|}{\multirow{6}{*}{ResNet34}}                                                              & \multicolumn{1}{c|}{\multirow{3}{*}{$10^{-4}$}} & (0.6, -0.06)                        & 0.812                   & 0.877                    \\
\multicolumn{1}{c|}{}                                                                                       & \multicolumn{1}{c|}{}                           & (0.8, -0.08)                        & 0.813                   & 0.874                    \\
\multicolumn{1}{c|}{}                                                                                       & \multicolumn{1}{c|}{}                           & (1, -0.1)                           & 0.814                   & 0.873                    \\ \cline{2-5} 
\multicolumn{1}{c|}{}                                                                                       & \multicolumn{1}{c|}{\multirow{3}{*}{$10^{-5}$}} & (0.6, -0.06)                        & 0.878                   & 0.877                    \\
\multicolumn{1}{c|}{}                                                                                       & \multicolumn{1}{c|}{}                           & (0.8, -0.08)                        & 0.878                   & 0.870                    \\
\multicolumn{1}{c|}{}                                                                                       & \multicolumn{1}{c|}{}                           & (1, -0.1)                           & 0.874                   & 0.871                    \\ \hline
\multicolumn{1}{c|}{\multirow{6}{*}{ResNet50}}                                                              & \multicolumn{1}{c|}{\multirow{3}{*}{$10^{-4}$}} & (0.6, -0.06)                        & 0.814                   & 0.887                    \\
\multicolumn{1}{c|}{}                                                                                       & \multicolumn{1}{c|}{}                           & (0.8, -0.08)                        & 0.814                   & 0.892                    \\
\multicolumn{1}{c|}{}                                                                                       & \multicolumn{1}{c|}{}                           & (1, -0.1)                           & 0.814                   & 0.886                    \\ \cline{2-5} 
\multicolumn{1}{c|}{}                                                                                       & \multicolumn{1}{c|}{\multirow{3}{*}{$10^{-5}$}} & (0.6, -0.06)                        & 0.860                   & 0.891                    \\
\multicolumn{1}{c|}{}                                                                                       & \multicolumn{1}{c|}{}                           & (0.8, -0.08)                        & 0.860                   & 0.888                    \\
\multicolumn{1}{c|}{}                                                                                       & \multicolumn{1}{c|}{}                           & (1, -0.1)                           & 0.853                   & 0.890                    \\ \hline
\end{longtable}
\endgroup

\begin{figure}[ht]
\centering
\includegraphics[width=1\textwidth]{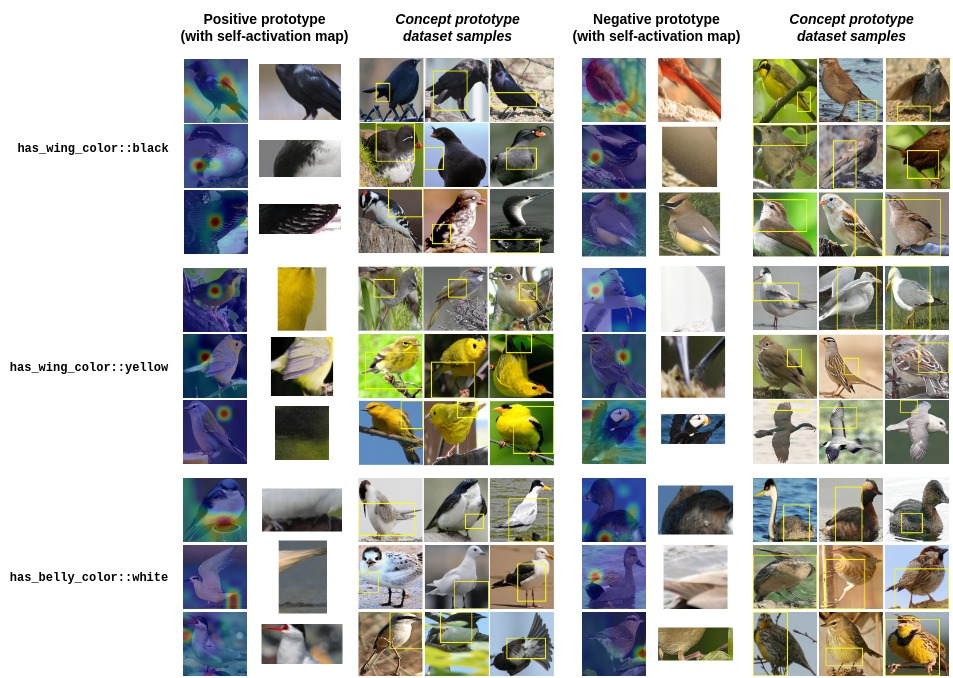}
\caption{Positive and negative prototypes with self-activation map and samples from the concept dataset (yellow square over the original image) for Concept-Guided ProtoPNet in the CUB dataset.}
\label{fig:ProtoPPNetCUB}
\end{figure}

\begin{figure}[ht]
\centering
\includegraphics[width=1\textwidth]{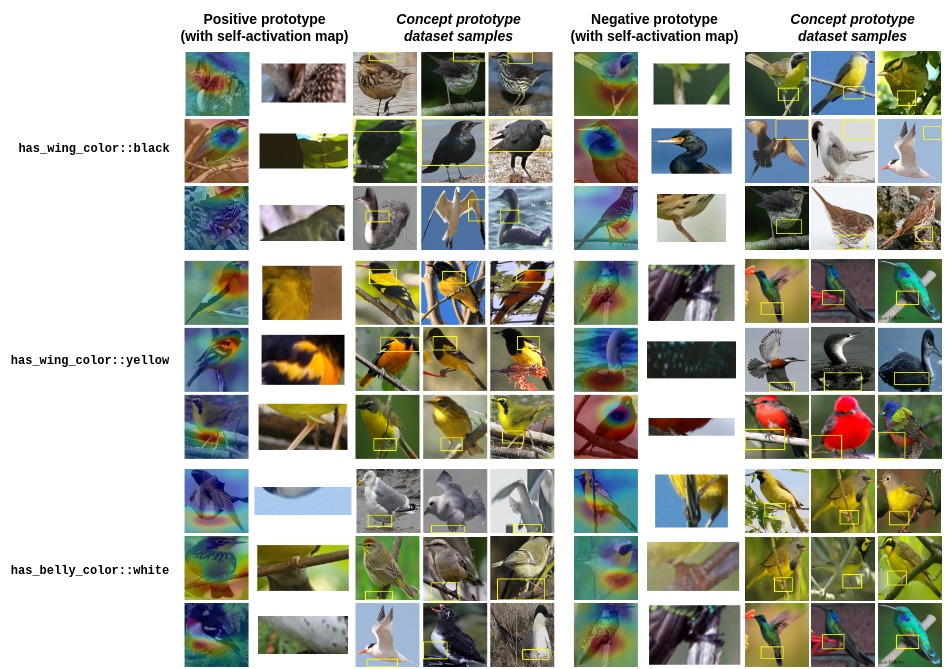}
\caption{Positive and negative prototypes with self-activation map and samples from the concept dataset (yellow square over the original image) for Concept-Guided ProtoPools in the CUB dataset.}
\label{fig:ProtoPPoolCUB}
\end{figure}

\newpage

\begin{figure}[ht]
\centering
\includegraphics[width=0.9\textwidth]{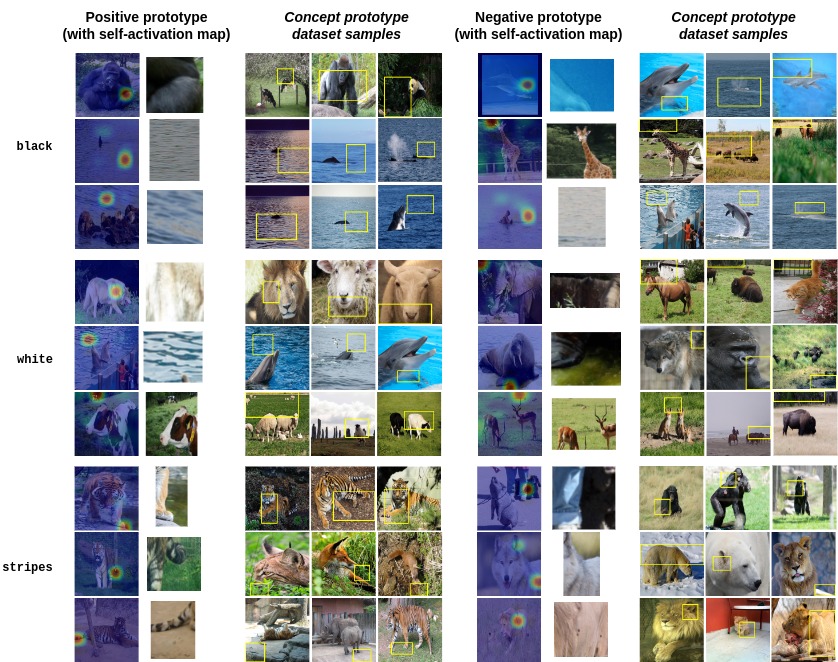}
\caption{Positive and negative prototypes with self-activation map and samples from the concept dataset (yellow square over the original image) for Concept-Guided ProtoPNet in the AwA2 dataset.}
\label{fig:ProtoPPNetAwA}
\end{figure}

\begin{figure}[ht]
\centering
\includegraphics[width=0.9\textwidth]{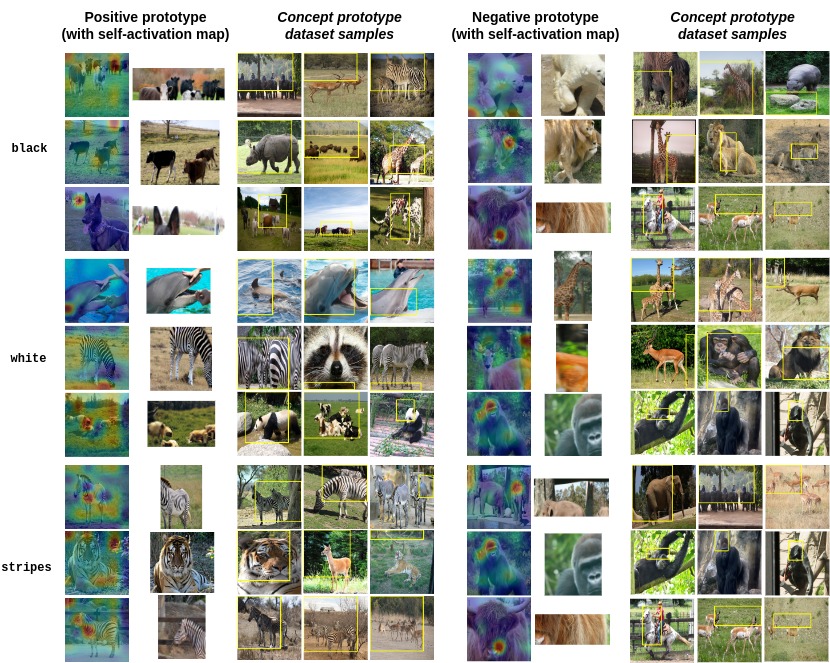}
\caption{Positive and negative prototypes with self-activation map and samples from the concept dataset (yellow square over the original image) for Concept-Guided ProtoPools in the AwA2 dataset.}
\label{fig:ProtoPPoolAwA}
\end{figure}


\end{document}